\documentclass[10pt,twocolumn,letterpaper]{article}
\usepackage{cvpr}
\usepackage{times}
\usepackage{epsfig}
\usepackage{graphicx}
\usepackage{amsmath}
\usepackage{amssymb}
\usepackage{multirow}
\usepackage[shortlabels]{enumitem}
\usepackage[pagebackref=true,breaklinks=true,letterpaper=true,colorlinks,bookmarks=false]
{hyperref}

\cvprfinalcopy

\def\cvprPaperID{4135} 

\begin{document}

\title{Dense xUnit Networks}

\author{Idan Kligvasser and Tomer Michaeli\\
Technion--Israel Institute of Technology,
Haifa, Israel\\
{\tt\small \{kligvasser@campus,tomer.m@ee\}.technion.ac.il}
}
\maketitle

\begin{abstract}
Deep net architectures have constantly evolved over the past few years, leading to significant advancements in a wide array of computer vision tasks. However, besides high accuracy, many applications also require a low computational load and limited memory footprint. To date, efficiency has typically been achieved either by architectural choices at the macro level (\eg using skip connections or pruning techniques) or modifications at the level of the individual layers (\eg using depth-wise convolutions or channel shuffle operations). Interestingly, much less attention has been devoted to the role of the activation functions in constructing efficient nets. Recently, Kligvasser \etal showed that incorporating spatial connections within the activation functions, enables a significant boost in performance in image restoration tasks, at any given budget of parameters. However, the effectiveness of their xUnit module has only been tested on simple small models, which are not characteristic of those used in high-level vision tasks. In this paper, we adopt and improve the xUnit activation, show how it can be incorporated into the DenseNet architecture, and illustrate its high effectiveness for classification and image restoration tasks alike. While the DenseNet architecture is extremely efficient to begin with, our dense xUnit net (DxNet) can typically achieve the same performance with far fewer parameters. For example, on ImageNet, our DxNet outperforms a ReLU-based DenseNet having $30\%$ more parameters and achieves state-of-the-art results for this budget of parameters. Furthermore, in denoising and super-resolution, DxNet significantly improves upon all existing lightweight solutions, including the xUnit-based nets of Kligvasser \etal.
\end{abstract} 
\section{Introduction}

\begin{figure}
\centering
\includegraphics[width=\columnwidth]{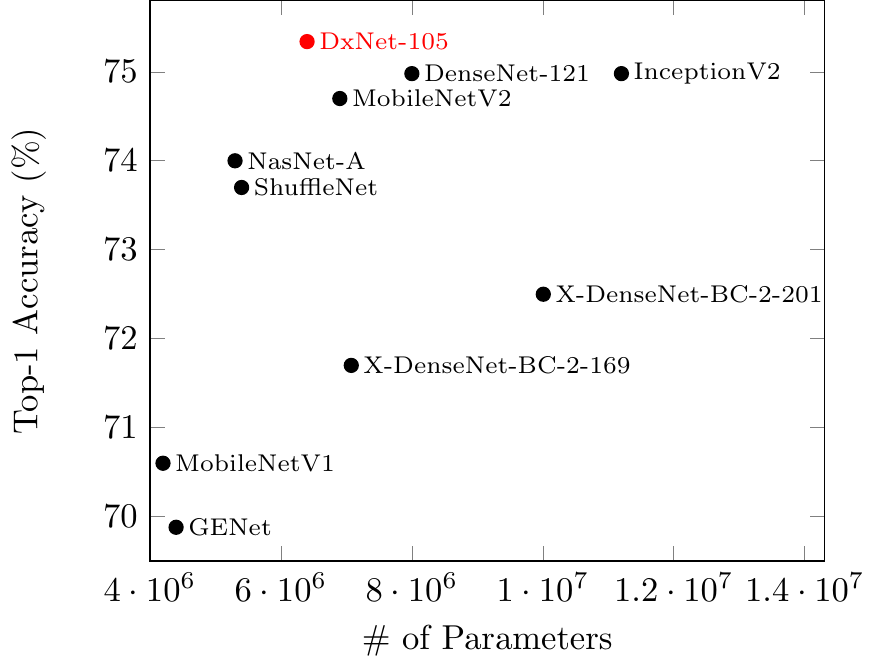}
\caption{\textbf{ImageNet classification performance of state-of-the-art lightweight models.} Top-1 accuracy rates ($224\times 224$ single-crop testing) on the ImageNet validation dataset as a function of the number of model parameters for our DxNet (red) in comparison with DenseNet \cite{densenets}, Inception-v2 \cite{inceptionv2}, MobileNet \cite{mobilenets}, MobileNet-v2 \cite{mobilenetv2}, ShuffleNet \cite{shufflenets}, GENet \cite{hu2018gather}, NasNet \cite{nasnets}, and two versions of Expander DenseNets (X-DenseNet-BC) \cite{prabhu2017deep}. See Sec.~\ref{sec:experiments} for details.}\label{fig:imagenet}
\end{figure}

Convolutional neural networks (CNNs) have led to significant advancements in computer vision, enabling to obtain extremely accurate results in high-level vision tasks such as classification \cite{vggs,resnets,densenets,mobilenets}, object detection \cite{yolo9000,fasterrcnn}, and image segmentation \cite{unet}, as well as in low-level vision tasks such as denoising \cite{dncnn,xunit} and super resolution \cite{srresnet,edsr}. The performance of CNN models has constantly improved over the last few years, a lot due to novel architectures that allow stable training of ever deeper nets \cite{resnets,densenets}. As a consequence, the impressive advancement in accuracy has been accompanied by a tremendous increase in memory footprint and computational burden. This trend has reached the point that in many computer vision tasks, the current state-of-the-art nets have hundreds of layers with tens of millions of parameters and are thus impractical for dissemination in mobile or wearable devices.

Many efforts have been invested in battling the large complexity of deep nets, mainly by proposing more effective architectures \cite{densenets,resnets,prabhu2017deep,iandola2016squeezenet}, pruning techniques \cite{lecun1990optimal,hassibi1993second,lebedev2016fast}, coefficient quantization \cite{hubara2016binarized,zhu2016trained,zhou2016dorefanet,rastegari2016xnor,hubara2016quantized}, and efficient convolution modules \cite{chollet2017xception,szegedy2017inception,mobilenets,mobilenetv2,shufflenets}. However, interestingly, in this ongoing quest for efficiency, much less attention has been devoted to the contribution of the nonlinear activation units. Recently, Kligvasser \etal~\cite{xunit} showed that incorporating spatial connections within the activation functions, can lead to a significant performance boost and often allows achieving state-of-the-art accuracy with much slimmer models. Yet this idea has only been explored in the context of image restoration (denoising, super-resolution, deraining) and within relatively small models ($\lesssim 1.5$M parameters), which are not characteristic of those used in classification.


In this paper, we explore the use of spatial activation units within DenseNets, which are architectures that already exhibit a very good tradeoff between accuracy and model size. As we show, our proposed dense xUnit nets (DxNets) provide a substantial further improvement in accuracy over conventional DenseNets, while requiring reduced model sizes. This allows us to obtain state-of-the-art results in classification as well as in image restoration, in the low-compute budget regime. For example, on the challenging ImageNet classification task, our DxNet outperforms the modern lightweight architectures \cite{densenets,inceptionv2,mobilenets,mobilenetv2,shufflenets,nasnets,prabhu2017deep} by a non-negligible margin in terms of both accuracy and number of parameters (see Fig.~\ref{fig:imagenet}).

In seek to decipher the mechanism that makes DxNets perform better, we find that upon convergence, these models are much less sensitive to parameter perturbations than their DenseNet counterparts. This suggests that the spatial connections within the activation functions cause the loss surface to be smoother, so that the optimization converges to a flatter minimum. As discussed in \cite{hochreiter1997flat,keskar2016large}, flatter minima tend to generalize better, thus suggesting a possible explanation for the improved performance of DxNets. We also find that the class activation maps \cite{zhou2016learning} of DxNets tend to capture the whole object in the image, while those of DenseNets often react to only portions of the object. These properties provide further support to the advantages of spatial nonlinearities observed in \cite{xunit}.



\section{Related work}
Stretching the trade-off between efficiency and accuracy in deep nets, has attracted significant research efforts over the last several years. Many solutions focus on the architecture at the macro level. 
The residual learning framework (ResNet) \cite{resnets} was the first to allow training of very deep nets, leading to a significant advancement in accuracy. For example, on the ImageNet dataset \cite{imagenet}, residual nets with a depth of up to 152 layers were trained, $8\times$ deeper than VGG nets \cite{vggs}. Later, the dense convolutional network (DenseNet) architecture \cite{densenets} took the ``skip-connections'' idea one step further, by connecting also distant layers within the net. This architecture ensures maximum information flow between layers and thus requires substantially less parameters to reach the same performance (\eg $50\%$ w.r.t.~ResNets). Further reduction in model sizes have been explored using pruning techniques \cite{lebedev2016fast,guo2016dynamic,liu2017learning} and via designs that ensure dense global connectivity with as little local connectivity as possible \cite{prabhu2017deep}.

Another family of techniques for constructing efficient nets focuses on improving the individual layers. In particular, many methods aim at making the convolution operations more efficient. In \cite{chollet2017xception}, a depth-wise convolution followed by a $1\times 1$ convolution was suggested as a replacement for the convolutional modules of Inception \cite{szegedy2017inception}. MobileNet~\cite{mobilenets,mobilenetv2} further utilized the depth-wise convolutions and demonstrated a significant reduction in size and latency, while preserving reasonable classification performance. The efficient ShuffleNet architecture \cite{shufflenets} reduced computation cost while maintaining accuracy by combining depth-wise convolutions with channel shuffle operations. Besides the convolution operations, efficiency can also be gained by activation function design. One notable example is the recently introduced xUnit \cite{xunit} module, which is a learnable nonlinear function with spatial connections. This module has been proposed as a replacement to the widespread per-pixel activations (\eg ReLUs). It has been illustrated that xUnits enable much more complex hidden representations, and thus lead to improved image-restoration results under any given computation budget.

Finally, we note in passing that many works focus on the level of machine computations, typically by using low-precision operations. These quantize the weights and activations of the net \cite{zhu2016trained,zhou2016dorefanet,rastegari2016xnor,hubara2016quantized}, so as to allow efficiet computations at both train and test times.

\section{Dense xUnit networks}

\begin{figure*}
\centering
\includegraphics[width=\textwidth]{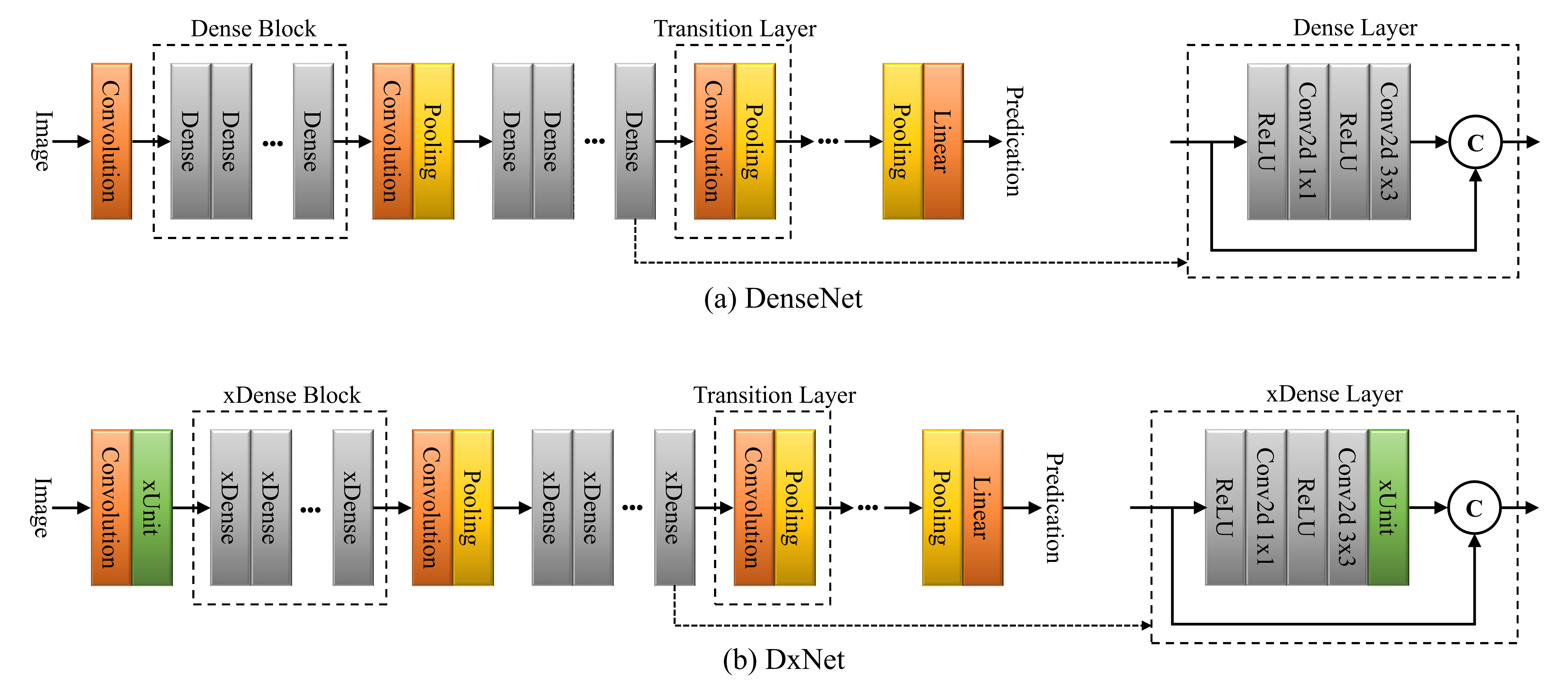}
\caption{\textbf{The dense xUnit network (DxNet) architecture.} (a) A DenseNet architecture comprises a sequences of dense-blocks and transition layers. Each dense-block consists of a sequence of dense layers. (b) The DxNet architecture utilizes xDense-blocks, which comprise a sequence of xDense-layers. These layers apply xUnits activations before performing the concatenation with their input feature maps.}\label{fig:scheme_densenet_dxnet}
\end{figure*}

At the architecture level, DenseNets \cite{densenets} provide one of the most efficient schemes in terms of accuracy for any given budget of parameters. However, this efficient macro-architecture, is commonly used with quite simple micro-blocks (\eg ReLU activations). Here, we propose to incorporate an xUnit-like activation function \cite{xunit} into DenseNets, as a means for further improving their efficiency. For simple feed-forward and ResNet models, xUnits can be used to replace all the ReLUs within the net. However, for DenseNets, such a strategy is impractical because of the significant overhead in parameters that accompanies the use of xUnits at layers with many feature maps. This motivates us to explore a different strategy. Rather than replacing the ReLU activations, we add xUnit activations at several locations along the net where the number of channels is small. This adds a relatively small number of learned parameters, but in return, has a dramatic effect on performance. Therefore, many layers can be discarded to maintain the same accuracy.

It is important to note that while hardware solutions can speed up computations at the level of a single layer, the operations of different layers cannot be parallelized due to their sequential nature. Therefore, on many platforms, achieving efficiency by means of discarding layers (as we do here) has a significant advantage over other approaches. We next provide a brief overview of DenseNets and xUnits, and then present our DxNets.

\subsection{DenseNets}
Dense convolutional networks (DenseNets) \cite{densenets} offer an extremely efficient macro-architecture, which comprises two key ingredients: Dense blocks and transition layers. A DenseNet is a stack of dense blocks followed by transition layers, as illustrated in Fig.~\ref{fig:scheme_densenet_dxnet}(a).

\textbf{Dense block:} A dense block consists of a sequence of ``dense layers'' having the form BatchNorm-ReLU-Conv(1x1)-BatchNorm-ReLU-Conv(3x3). Each such dense layer outputs a fixed number $k$ of feature maps, which are concatenated to its input maps and passed on to the next dense layer. Thus the number of feature maps after each dense layer increases by $k$, which is called the growth rate of the net. This implies that a dense block with $n$ dense layers, outputs $n\times k$ more feature maps than the number of feature maps at its input.

\textbf{Transition layer:} The transition layer serves to squeeze back the representation after each dense block, both in terms of the number of feature maps and in terms of their spatial support. If the number of feature maps at the input of a transition map is $m$, then it generates $r\times m$ output feature maps, where the parameter $r<1$ is called the reduction rate of the DenseNet. More concretely, the transition layer has the form BatchNorm-Conv(1x1)-AveragePooling(2x2).

While traditional feed-forward architectures invest a significant amount of parameters in preserving the current state within the net, DenseNet is a very slim architecture because it explicitly differentiates between information that is added and information that is preserved. That is, each dense block adds a limited set of feature-maps while keeping the remaining feature-maps unchanged. Thus, the DenseNet architecture requires fewer parameters and less computation than traditional convolutional networks to achieve the same performance.

In addition to parameter efficiency, the DenseNet architecture also mitigates the vanishing-gradient problem by allowing improved information flow. Due to the dense connections, each layer has access to the gradients from the loss function. The result is that, as shown in \cite{densenets}, DenseNet models are very easy to train.

\subsection{xUnit activations}

\begin{figure}
\centering
\includegraphics[width=0.7\columnwidth]{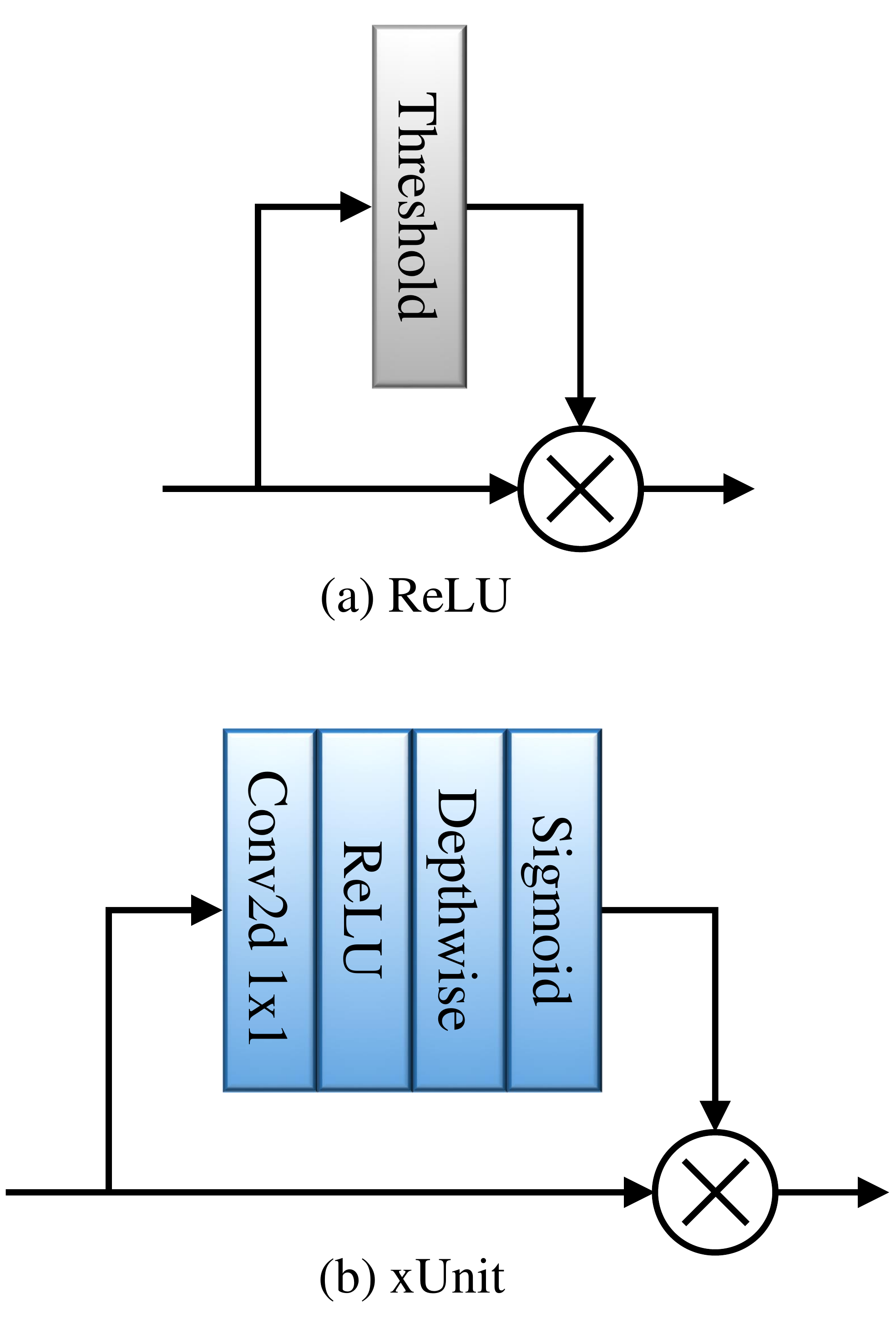}
\caption{\textbf{The xUnit activation layer.} (a) The ReLU activation function can be interpreted as multiplying the input by a binary weight map, which is constructed by applying an element-wise threshold on the input. (b) The xUnit activation constructs a weight map in the range $[0,1]$ by applying a nonlinear \emph{spatial} operation on the input.}\label{fig:scheme_relu_xunit}
\end{figure}

The recently suggested xUnit \cite{xunit} is a learnable nonlinear activation function with spatial connections. Specifically, most popular activation functions operate element-wise on their arguments, and can be interpreted as performing a Hadamard product between their input and a weight map. For example, the popular ReLU \cite{relu} multiples its input by a binary weight map, which is a thresholded version of the input. This is illustrated in Fig.~\ref{fig:scheme_relu_xunit}(a). By contrast, the authors of \cite{xunit} suggest to use ``learnable spatial activations''. The basic idea is to construct a weight map in which each element depends on the spatial neighborhood of the corresponding input element. This weight map is constructed by passing the input through a ReLU followed by a depth-wise convolution, and some element-wise gating function which maps the dynamic range to $[0,1]$ (\cite{xunit} used a Gaussian, we use a sigmoid). Each nonlinearity within the xUnit is preceded by batch normalization.

As each xUnit introduces an additional set of learnable parameters, merely replacing ReLUs by xUnits clearly increases memory consumption and computation. However, xUnits offer a performance boost, thus allowing to use a smaller number of layers to achieve the same performance. The authors of \cite{xunit} show empirically that the optimal percentage of parameters to be invested in the activations is at least $15\%$ for simple feed-forward models. This is while nets with per-element nonlinearities invest $0\%$ of the parameters in the activations. As an example, replacing ReLUs by xUnits in a conventional denoising model has been reported in \cite{xunit} to allow reduction of the number of parameters by $2/3$ while achieving the same restoration accuracy.

\begin{figure}
\includegraphics[width=0.9\columnwidth]{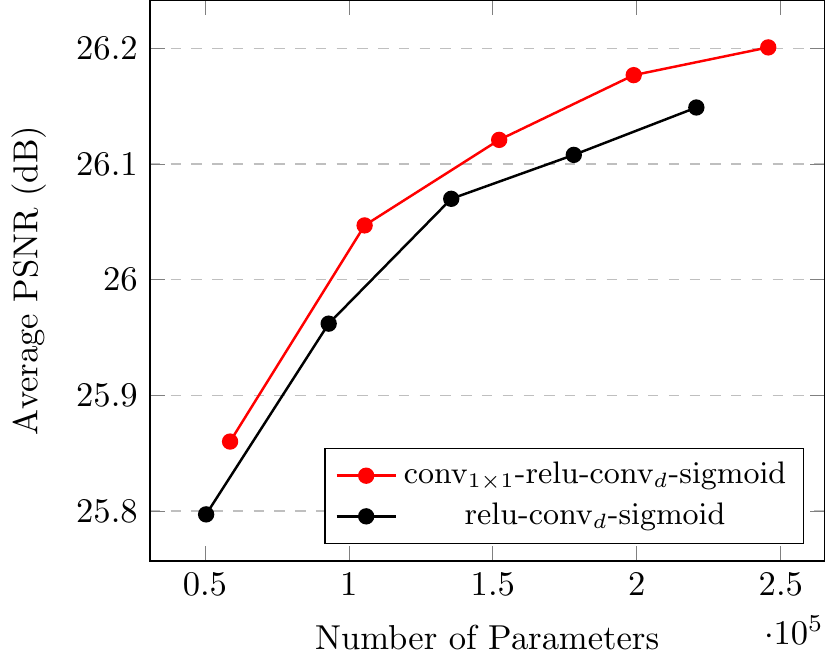}
\caption{\textbf{The impact of cross-channel dependencies in xUnits.} The graph compares the denoising performance of two xUnit designs, incorporated within a conventional ConvNet architecture (Conv+xUnit layers). We gradually increase the number of layers and record the average PSNR obtained in denosing the BSD68 dataset with noise level of $\sigma = 50$. Training configurations are the same for both nets. Our design, which uses a $1\times 1$ convolutional layer, achieves a better tradeoff between performance and model size, compared to the original xUnit design of \cite{xunit}.} \label{fig:xunit_var}
\end{figure}

Here, we introduce an important modification to the xUnit design proposed in \cite{xunit}, which is the addition of a $1\times 1$ convolutional layer at the input of the branch constructing the weight map. Our xUnit design is shown in Fig.~\ref{fig:scheme_relu_xunit}(b). The $1\times 1$ convolution layer allows cross feature map dependencies, and improves performance. This is illustrated in Fig.~\ref{fig:xunit_var} for the task of denoising with a simple feed-forward net. For the depth-wise convolution within the xUnit, we always use $9\times 9$ kernels, as suggested in \cite{xunit}.

\subsection{Suggested network architecture}
To incorporate xUnits into DenseNets, we use the fact that although the number of feature maps may become quite large along the net, the growth rate $k$ is typically modest (\eg no larger than $32$). Therefore, inside each dense block, we have a slim bottleneck, right before the concatenation with the input feature maps. Those paths are ideal for adding units with learnable parameters, as the resulting overhead in model size is relatively modest. Specifically, an xUnit operating on a $k$-channel feature map, contributes only $k\times 9^2 + k= 82 k$ parameters to the model. In return to this modest overhead, our experiments indicate that a substantial improvement in performance can be obtained. Therefore, as illustrated in Fig.~\ref{fig:imagenet}, for example, this allows to achieve a reduction in the total number of parameters (by discarding layers) while maintaining the same performance. We call our modified dense layers x-dense layers; a sequence of such layers form an x-dense block; and the overall resulting architecture is coined a \textit{Dense xUnit Network} (DxNet). This architecture is depicted in Fig.~\ref{fig:scheme_densenet_dxnet}(b).

\section{Experiments}\label{sec:experiments}
We examine DxNet's performance in several high and low level vision tasks.

\subsection{Image classification}

\begin{figure*}
\begin{center}
\includegraphics[width=\textwidth]{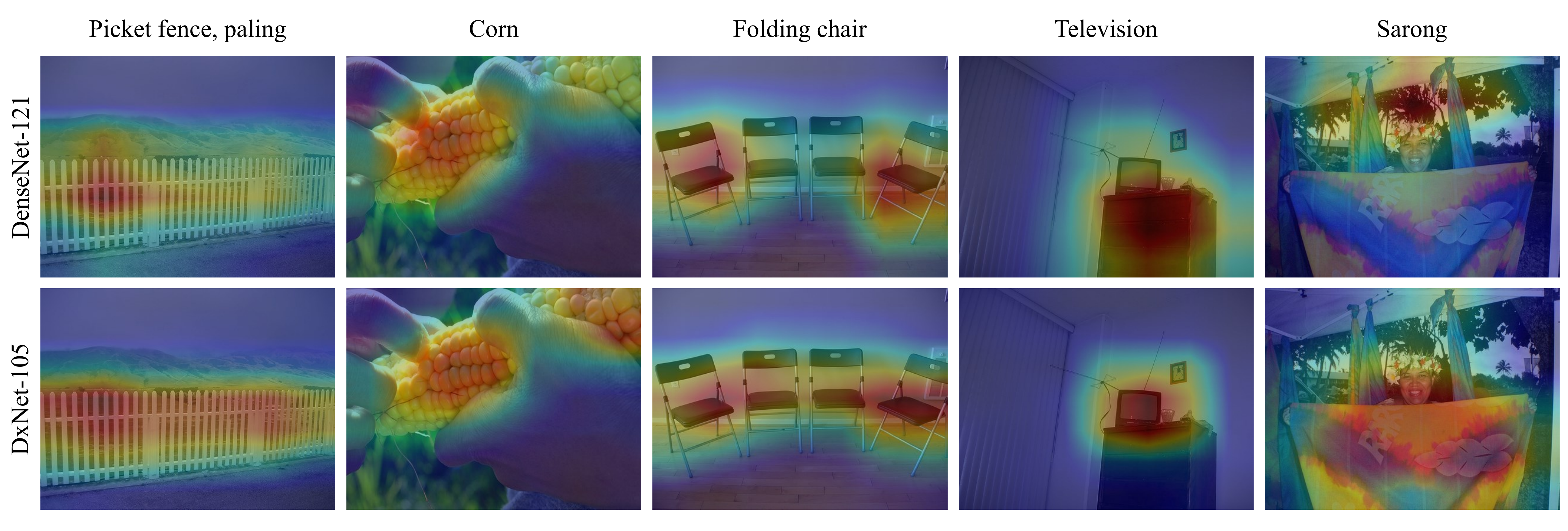}
\end{center}
\caption{\textbf{Class activation maps.} Example CAMs for several images, obtained with DenseNet (top) and with our DxNet (bottom). The maps highlight the image regions used for classification.}\label{fig:cam_visualization}
\end{figure*}

We begin by illustrating the effectiveness of DxNets in three image classification tasks: CIFAR-10 \cite{cifar}, SVHN \cite{svhn} and ILSVRC 2012 \cite{imagenet}. In all our experiments, we use stochastic gradient descent (SGD) optimization with a weight decay of $5\cdot10^{-4}$ and Nestrov momentum \cite{nesterov} of 0.9. The learning rate is initialized to $10^{-1}$ and is reduced by a factor of $2$ when the validation loss stops improving (reduce on plateau). The exact network configurations for each task are briefly summarized below and thoroughly detailed in the Supplementary Material.

\paragraph{CIFAR-10.} The CIFAR-10 is a traditional computer-vision dataset, which is used for developing object recognition models. It consists of $60,000$ $32\times 32$ color images, containing 10 object classes, with 6000 images per class. Here we use a network with three dense blocks, each with the same number of dense layers (see Table~\ref{tab:cifar}). The first dense block is preceded by a convolution with $24$ output channels, followed by an xUnit activation. We perform $2\times 2$ average pooling between every two adjacent dense blocks. At the end of the last dense block, a global average is applied, followed by a softmax classifier. We set the growth rate to $k=12$ and the reduction rate to $r=0.5$. As models with high capacity are likely to suffer from over-fitting on this relatively small dataset, we adopt a data augmentation technique from \cite{cutout}. Specifically, we employ normalization using per-channel mean and standard deviation, random cropping, random horizontal mirroring and random cutout of $16\times 16$ patches. We run the optimization for 200 epochs and use a batch-size of 128. Table \ref{tab:cifar} compares the suggested DxNet to the state-of-the-art ReLU based DenseNet \cite{densenets} architecture. Since our data augmentation procedure is different from that of \cite{densenets}, here we retrained the DenseNet models using the same augmentation strategy we used for our DxNets. These DenseNet models achieve improved results w.r.t.\@ those reported in \cite{densenets}. For example, a DenseNet with $k=12$ and a configuration of $16-16-16$ achieves an error rate of $4.03\%$ with the cutout augmentation, compared to the $4.51\%$ reported in \cite{densenets}. Yet, as can be seen in the table, our DxNet models can achieve roughly the same error rates with significantly less network parameters.

\paragraph{SVHN.} The Street View House Numbers (SVHN) is a real-world image dataset obtained from house number images. The SVHN contains $600,000$ $32\times 32$ color images. Here we use the same models as in the CIFAR-10 experiment. Following \cite{densenets}, we add a dropout layer with a rate of 0.2 after each convolutional layer and do not use any data augmentation. We run the optimization for 100 epochs and use a batch size of 128. As can be seen in Table \ref{tab:cifar}, in this task as well, our DxNets match the performance of DenseNets with far fewer parameters.

\paragraph{ImageNet.} The ILSVRC 2012 is a large visual dataset, which contains $1.2$ million images for training, and $50,000$ for validation, within $1,000$ categories. Here we use a dense block configuration of 6-12-20-12, and growth and reduction rates of $k=32$ and $r=0.5$, respectively. We use the same data augmentation as in the DenseNet \cite{densenets} training procedure. We train our models with a batch-size of 256 for 90 epochs. Figure~\ref{fig:imagenet} compares our model against the state-of-the-art lightweight models DenseNet-121 \cite{densenets}, InceptionV2 \cite{inceptionv2}, MobileNetV1 \cite{mobilenets}, MobileNetV2 \cite{mobilenetv2}, ShuffleNet \cite{shufflenets}, X-DenseNet \cite{prabhu2017deep}, GENet \cite{hu2018gather} and NasNet-A \cite{nasnets}. This figure depicts the top-1 classification accuracy vs.\@ the number of network parameters for single-crop $224\times 224$ images on the validation dataset. As can been seen, our DxNet-105 outperforms all state-of-the-art light-weight models in terms of classification accuracy, while being very slim in terms of number of parameters. In order to better understand the effect of the spatial nonlinearities, we show in Fig.~\ref{fig:cam_visualization} class activation map (CAM) \cite{zhou2016learning} visualizations, which highlight the image regions used by the nets to classify the category. For fair comparison we choose images from the validation set, which both nets classify correctly. Our DxNet tends to react to regions that are tightly supported on the entire object or on its discriminative parts. This is while DenseNet often resonates to only portions of the object and to some of the background. Please see the Supplementary Material for many more visualizations.

\begin{table}[t]
\begin{center}
\begin{tabular}{ |c|c c c|c|c|c| }
\hline
Method  & Configuration & Params & C10 & SVHN \\
\hline
\hline
DenseNet & 16-16-16 & 0.8M & 4.03 & 1.76 \\
DxNet & 12-12-12 & 0.5M & 4.13 & 1.78 \\
\hline
DenseNet & 26-26-26 & 1.7M & 3.83 & 1.74 \\
DxNet & 19-19-19 & 1.0M & 3.80 & 1.73 \\
\hline
\end{tabular}
\end{center}
\caption{\textbf{Classification performance.} Error rates (\%) on the CIFAR-10 and SVHN datasets. Our DxNets achieve roughly the same performance as DenseNets, but with far fewer parameters.}\label{tab:cifar}
\end{table}

\subsection{Single image super resolution}
Our DxNet architecture can also be applied in single image super resolution. We specifically focus on $4\times$ super-resolution for images down-sampled with a bicubic kernel. Influenced by the EDSR framework \cite{edsr}, for this task we remove all batch normalization layers and use mean-absolute error as our training loss. All our models are trained using the Adam optimizer \cite{adam} with its default settings. The learning rate is initialized to $10^{-4}$ and reduced to $10^{-5}$ at $50\%$ of the total number of training epochs. The mini-batch size is set to 16 for 6000 epochs. We use 800 2K resolution images from the DIV2K dataset \cite{ntire2017}, enriched by random cropping and 90$^\circ$ rotations.

Here, we examine several different version of the DxNet architecture. The first is a narrow network with a growth rate of $k=16$ and dense block configuration 4-4-4-6-8-8-8, which we coin DxNet$_{16}$. The second, which we coin DxNet$_{32}$, is a wider network with a growth rate of $k=32$ and dense block configuration of 4-6-8. We compare our models with several state-of-the-art lightweight super-resolution methods, including VDSR \cite{vdsr}, LapSRN \cite{lapsrn}, MS-LapSRN \cite{lai2018fast} MemNet \cite{memnet}, SRResNet \cite{srresnet}, CARN \cite{carn}, SRDenseNet \cite{srdensenet}, ProSR \cite{prosr} and EDSR \cite{edsr} (baseline version). The comparison is based on measuring the PSNR on the y-channel while ignoring a 4 pixels frame at the border of the reconstructed images.

\begin{figure}
\centering
\includegraphics[width=\columnwidth]{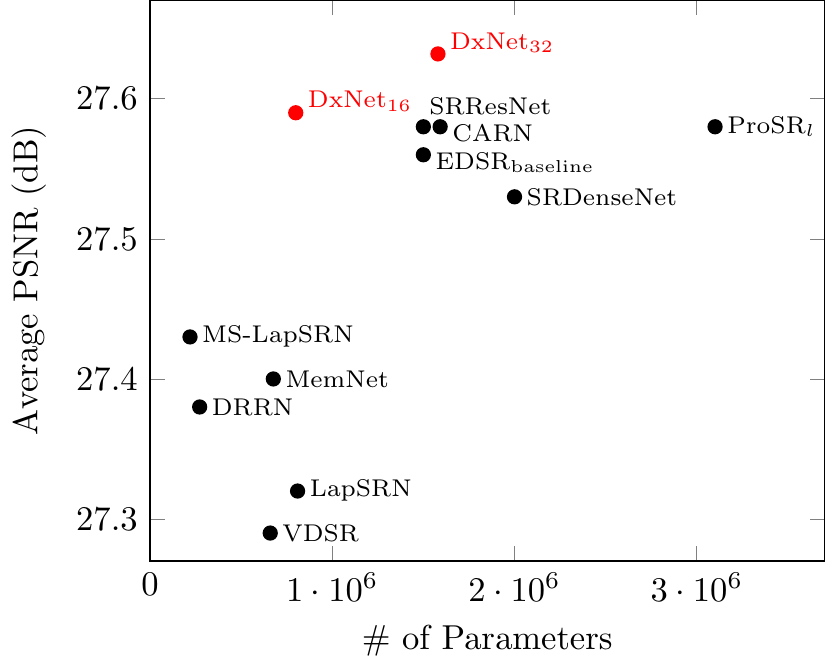}
\caption{\textbf{Super resolution performance vs.\@ number of parameters.} The average PSNR in [dB] attained in the task of 4x SR on the BSD100 dataset. Our DxNets attain a significantly higher PSNR at any given budget of parameters.}\label{fig:sr_psnr_vs_params}
\end{figure}

Figure \ref{fig:sr_psnr_vs_params} depicts the PSNR attained by all models on the BSD100 dataset \cite{bsd}. Our models achieve state-of-the-art results, while being much smaller than the competing architectures. In particular, our 0.86M parameter DxNet$_{16}$ remarkably achieves a PSNR score which is on par with the 3.1M parameter DenseNet based model, ProSR$_{l}$ \cite{prosr}. This corresponds to nearly a 4-fold reduction in model size.

\begin{figure*}
\begin{center}
\includegraphics[width=\textwidth]{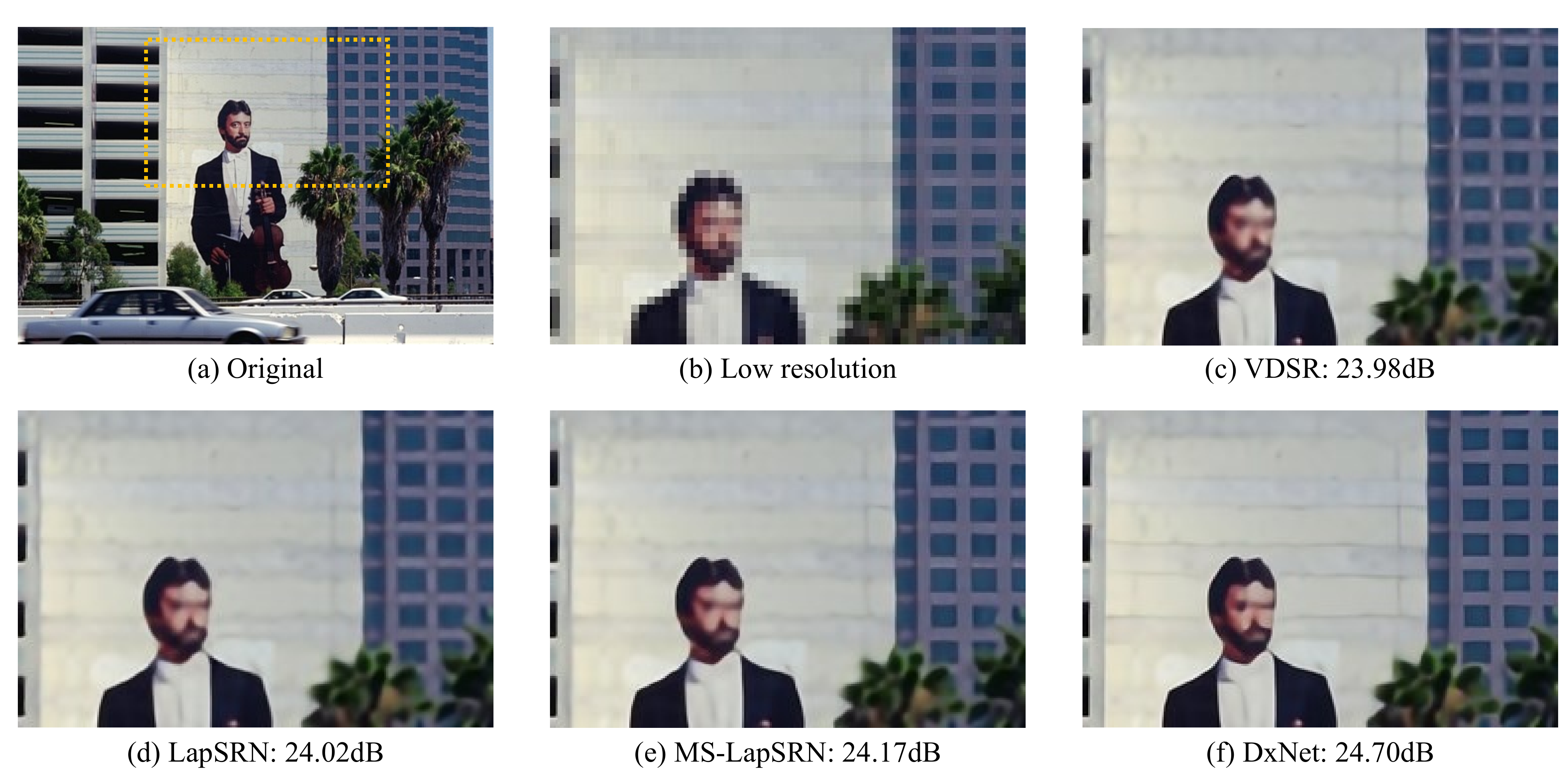}
\end{center}
\caption{\textbf{Image super resolution result for magnification $\times 4$.} Comparison of the reconstructed images produced by VDSR, LapSRN, MS-LapSRN and our DxNet$_{16}$ for magnification of $\times 4$. All models contain less than a million of network parameters. In contrast to the competing methods, our DxNet$_{16}$ manages to restore more of the image details.}\label{fig:sr_visualization}
\end{figure*}

Figure \ref{fig:sr_visualization} shows example $4\times$ super resolution results obtained with our DxNet$_{16}$, as well as with the state-of-the-art slim models VDSR, LapSRN, and MS-LapSRN (all having less than $1$M parameters). As can been seen, our DxNet$_{16}$ manages to restore more of the fine image details.

\subsection{Image denoising}\label{exp:densoing}
Next, we illustrate the efficiency of our architecture in image denoising. In this task, the size of the output image is the same as that of the input image, therefore, we remove all pooling and fully connected layers. We study the DxNet architecture with growth rate $k=16$ and reduction rate $r=0.5$.

All networks are trained using the Adam optimizer \cite{adam} with its default settings and with the mean squared error (MSE) loss. We train for 5000 epochs with batches of 32 images. The learning rate is initially set to $10^{-3}$, and scheduled to decrease by a factor of $5$ at $10\%$, $25\%$, $75\%$ and $90\%$ of the total number of epochs. We use 400 images from the BSD dataset \cite{bsd}, augmented by random horizontal flipping and cropping. The noisy images are generated by adding Gaussian noise to the training images. We use residual learning; that is, the net is trained to predict the noise and this noise estimate is then subtracted from the noisy image at test time.

\begin{figure}
\centering
\includegraphics[width=\columnwidth]{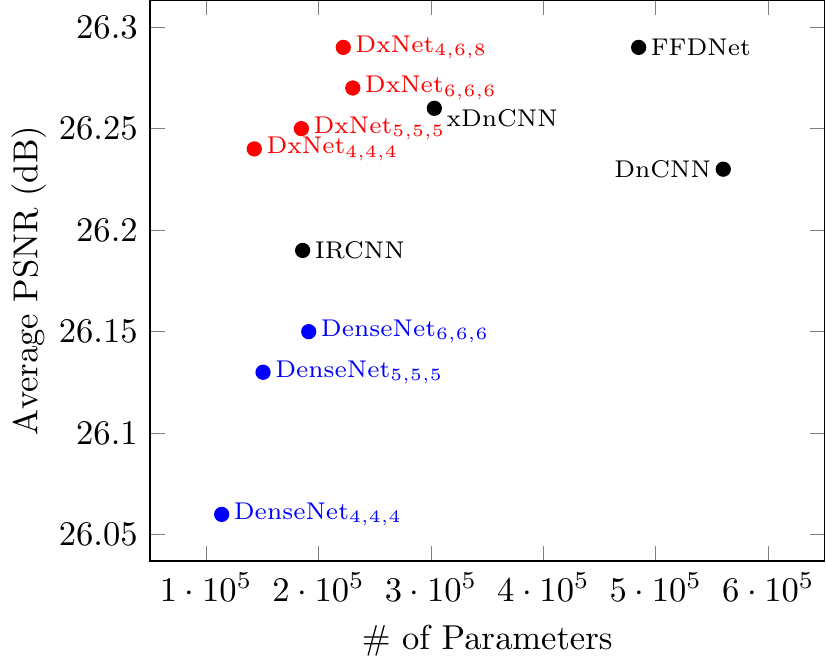}
\caption{\textbf{Denoising performance vs.\@ number of parameters.} We compare several methods with our DxNets in the task of denoising with noise level $\sigma = 50$. Our DxNets have the best tradeoff between performance and efficiency.}\label{fig:denoising_psnr_vs_params}
\end{figure}

Figure \ref{fig:denoising_psnr_vs_params} compares the average PSNR attained by our DxNets to that attained by the original DenseNets, as well as to the state-of-the-art (non-dense) methods xDnCNN \cite{xunit}, DnCNN \cite{dncnn} IRCNN \cite{zhang2017learning} and FFDNet \cite{zhang2018ffdnet}. The evaluation is performed on the BSD68 dataset, a subset of 68 images from the BSD dataset \cite{bsd}, which is not included in the training set. As can be seen, our DxNet models outperform all other denoising methods, while being significantly smaller. For example, our DxNet$_{4,4,4}$ (having a dense block structure of 4-4-4), performs slightly better than DnCNN, although its size is one fourth of DnCNN in terms of number of parameters.

\subsection{Minima sharpness}
To understand how the incorporation of xUnit activations affects the loss surface during training, it is insightful to examine the sensitivity of the net to small perturbations in its parameters. At a local minimum, such an analysis provides indication for the flatness of the minimum. This is thus informative since, as shown in \cite{hochreiter1997flat,keskar2016large}, flatter minima tend to generalize better.

\begin{figure}
\includegraphics[width=0.88\columnwidth]{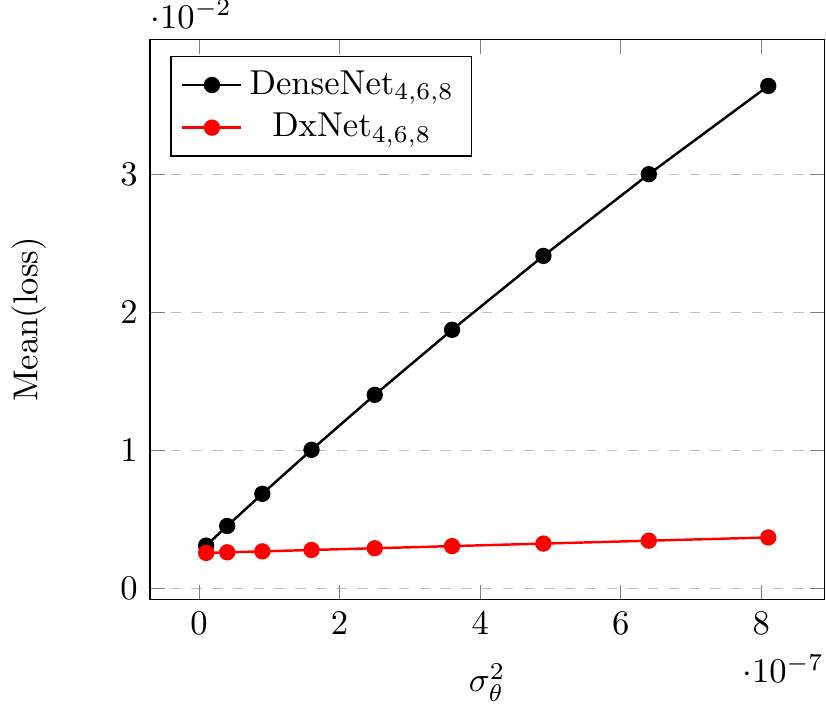}
\caption{\textbf{Loss vs.\@ parameter perturbation.} We compare between DenseNet and DxNet in a denoising task, in terms of their resilience to Gaussian perturbations of their parameters. Both networks have a similar structure of 4-6-8 blocks. We gradually increase the standard deviation of the noise, which we add to all convolutional filters, and record the average mean square error obtained in denoising the BSD training dataset with noise level $\sigma = 50$ for 1000 different permutations. The training configurations are the same for both of networks. Our DxNet is substantially less sensitive to parameter perturbations, a property which has been linked to better generalization.}\label{fig:stability_mean}
\end{figure}

\begin{figure}
\includegraphics[width=0.88\columnwidth]{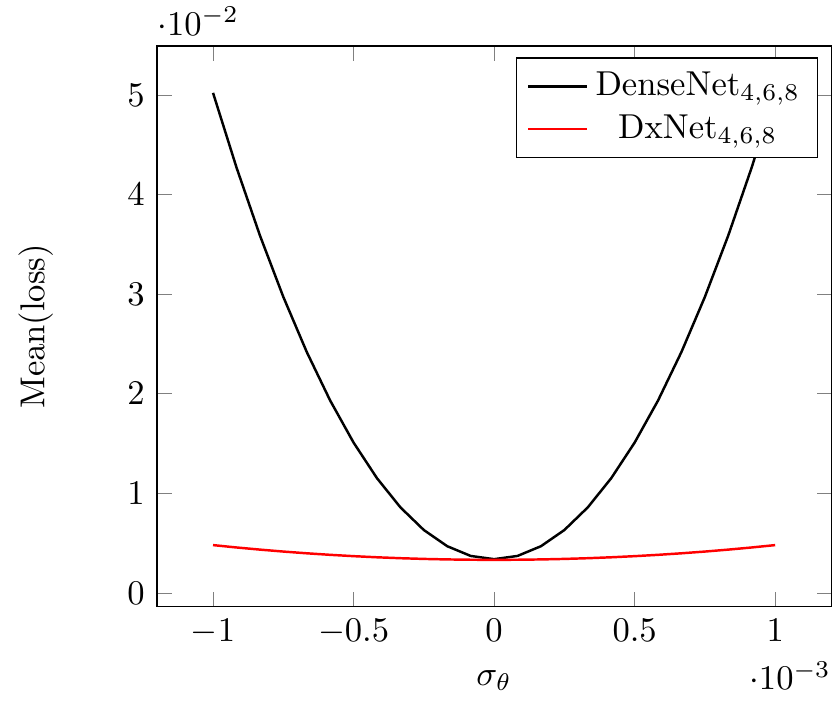}
\caption{\textbf{Visualization of loss around the minimum.} We visualize the loss around the minima that DenseNet and DxNet converged to in the experiment of Fig.~\ref{fig:stability_mean}, by showing 1D quadratic functions whose second order derivatives match the Hessians of the corresponding high-dimensional loss surfaces. This illustration highlights that our DxNet loss surface is significantly flatter around the minimum, a property which has been linked to better generalization.}\label{fig:est_loss}
\end{figure}

Consider a model $f_\theta$, having parameters $\theta$, which takes an input $x$ and outputs a prediction $y$. The model is trained to minimize some loss function, $L(\theta)$, which measures the average discrepancy between the ground-truth and predicted labels over the training set. Around a local minimum point $\theta_0$, a second-order Taylor expansion of $L(\theta)$ takes the form
\begin{align}
L(\theta) \approx L(\theta _0)+ \frac{1}{2} (\theta-\theta _0)^T H_L (\theta _0) (\theta-\theta _0),
\end{align}
where we used the fact that the gradient vanishes at $\theta_0$, and denoted the Hessian by $H_L$.
Now, if we randomly draw $\theta$ from a spherical Gaussian distribution around $\theta_0$, as $\theta\sim\mathcal{N}(\theta_0,\sigma_\theta^2 I)$, then we get that the mean loss of the model with the perturbed parameters is given by
\begin{equation}\label{eq:mean_loss}\mathbb{E} [L(\theta)] = L(\theta _0)+\frac{1}{2} \text{Tr}(H_L) \sigma^{2}_{\theta}.
\end{equation}
That is, the average loss of the perturbed model (calculated over many realizations of perturbed parameters) is linearly proportional to the perturbation strength $\sigma^2_{\theta}$, with a slope of $\frac{1}{2}\text{Tr}(H_L)$. This observation can be used to estimate $\text{Tr}(H_L)$, which can be thought of as a measure of the flatness of the minimum. This is illustrated in Fig.~\ref{fig:stability_mean}. Here, we trained two image denoising nets, as in Sec.~\ref{exp:densoing}. Both networks have a block structure of 4-6-8. One is a plain DenseNet (which we coin DenseNet$_{4,6,8}$), and one is a DxNet (coined DxNet$_{4,6,8}$). We gradually increased the standard deviation $\sigma_\theta$ of the perturbation noise which we add to all convolutional filters of the trained models, and recorded the MSE obtained in denoising the training dataset for 1000 noise realizations. As can be seen, the losses of the perturbed models indeed increase (roughly) linearly with $\sigma_\theta$. Our DxNet$_{4,6,8}$ is far more resilient to parameter perturbation. The mean of the Hessian eigenvalues (calculated from the slopes of the two graphs) is 0.512 for DenseNet and 0.013 for our DxNet. This indicates that the minimum of the loss for our model is much flatter. Figure \ref{fig:est_loss} visualizes this difference by plotting 1D quadratic functions whose second-order derivatives match the Hessians of the high-dimensional loss surfaces.

\section{Conclusion}
In this paper, we introduced DxNet, a deep net architecture that allows attaining the same accuracy as DenseNets, but with far smaller model sizes (having less layers and less parameters). Our DxNet is based on the DenseNet architecture, with the addition of nonlinear activation functions that have spatial connections. These allow the construction of stronger features without a significant increase in the number of net parameters. Therefore, as we illustrated on several classification and image restoration tasks, our DxNet typically outperforms DenseNet by a large margin under any given budget of parameters. We also showed that DxNet is less sensitive to perturbations in its parameters than DenseNet, a property that has been previously linked to better generalization. We believe these properties make DxNet highly suited to a wide range of Computer Vision tasks.

\pagebreak

{\small
\bibliography{bib}
\bibliographystyle{ieee}
}

\end{document}